\documentclass[12pt]{amsart}

\usepackage{amsrefs}
\usepackage[english]{babel}
\usepackage[utf8x]{inputenc}
\usepackage{geometry} 
\usepackage{euscript}
\usepackage{graphicx} 
\usepackage{epstopdf}
\usepackage{amscd}
\usepackage{array} 
\usepackage{verbatim} 
\usepackage{amssymb} 
\usepackage{amsmath}
\usepackage{amsthm}

\newtheorem{theorem}{Theorem}
\newtheorem{lemma}[theorem]{Lemma}

\theoremstyle{definition}

\theoremstyle{remark}

\theoremstyle{remark}

\theoremstyle{remark}

\theoremstyle{remark}

\renewcommand{\phi}{\varphi}

\begin{document}

\title[Affine neural networks]{Approximation of functions by neural networks}

\author{Andreas Thom}
\address{Andreas Thom, TU Dresden, Germany}
\email{andreas.thom@tu-dresden.de}


\begin{abstract}
We study the approximation of measurable  functions on the hypercube by functions arising from affine neural networks. Our main achievement is an approximation of any measurable function $f \colon W_n \to [-1,1]$ up to a prescribed precision $\varepsilon>0$ by a bounded number of neurons, depending only on $\varepsilon$ and \emph{not} on the function $f$ or $n \in \mathbb N$.
\end{abstract}

\maketitle



The study of functions defined by neural networks has a long history dating back to the work of McCulloch and Pitts \cite{mp}. Recent advances in applications to deep learning raised numerous questions on why neural networks are able to solve so many different problems. It is well known that neural networks can approximate any given function up to arbitrary precision, see for example \cite{a1,a2,a3}, however, the dependence of the architecture of the neural network on the quality of the approximation and the function is much harder to understand on a theoretical level. This is in contrast to the observation that relatively easy neural networks are able to provide desirable approximations in many cases. Our take on this is a new viewpoint that tries to approach this phenomenon as an effect of efficient separation of structure and randomness.

\vspace{0.2cm}

We consider the standard hypercube $W_n := [-1,1]^n \subset \mathbb R^n$ and an arbitrary parameter $q\geq1$. Every vector $\xi=(\xi_0,c)\in [-q,q]^{n} \times \mathbb R$ naturally defines a function 
$$\varphi_\xi \colon W_n \to [-1,1], \quad \varphi_\xi(w) := \beta(\langle w,\xi_0 \rangle + c),$$ 
where $$\beta(z)= \begin{cases} 1 & z \geq 1 \\
z & z \in [-1,1] \\
-1 & z \leq -1 \end{cases}.$$
and $\langle.,.\rangle$ denotes the standard inner product on $\mathbb R^n$. We call such functions \emph{rectified affine} and suppress the parameter $q$ throughout the entire article.
In this note, we consider the basic question how easily an arbitrary measurable function $f \colon W_n \to [-1,1]$ can be approximated by compositions and insertions of functions of the form $\varphi_\xi$ as above. Our main achievement is an approximation of any measurable function $f \colon W_n \to [-1,1]$ up to a prescribed precision $\varepsilon>0$ by a bounded number of neurons, where the bound depends only on $\varepsilon$ and \emph{not} on $n \in \mathbb N$.
This independence of $n \in \mathbb N$ is reminiscent of Szemer\'edi's Regularity Lemma, which provides a decomposition of a graph of arbitrary size into quasi-random and deterministic constituents. In a similar way, we decompose an arbitrary measurable function $f \colon W_n \to [-1,1]$ into constituents that can be reproduced by a neural network of bounded size and such which are quasi-random and thus invisible to a neural network of bounded size in a very precise way.

Let us introduce a hierarchy of functions built from the basic building blocks. A function $f \colon W_n \to [-1,1]$ is said to be represented by an affine neural network of type $(d_1,d_2,...,d_r)$, where $r \in \mathbb N$, if there exists a sequence
$$W_n \stackrel{\alpha_0} \to W_{d_1} \stackrel{\alpha_1}{\to} \cdots \stackrel{\alpha_{r-1}}{\to} W_{d_r}  \stackrel{\alpha_r}{\to} W_1= [-1,1],$$
where we set $d_0=n$ and $d_{r+1}=1$ for convenience and each $\alpha_0,\dots,\alpha_n$ is
of the form $$\alpha_i = (\varphi_{\xi(i,1)},\dots,\varphi_{\xi(i,d_{i+1})}), \quad 1 \leq i \leq r$$
with $\xi(i,j) \in [-q,q]^{d_i} \times \mathbb R$, for $1 \leq j \leq d_{i+1}$ We say that $f \colon W_n \to [-1,1]$ is $(d|r)$-representable if it is representable by an affine neural network of type $(d_1,\dots,d_r)$ with $d_i \leq d$ for all $1 \leq i \leq r$.
A function is $(d|0)$-representable function if and only if it is an rectified affine function. It is also easy to see that $(d|r)$-representable functions are both $(d+1|r)$-representable and $(d|r+1)$-representable. In particular, any rectified affine function is $(d|r)$-representable for any $d \geq 1, r\geq 0$.

\begin{lemma} \label{para}
Let $f_i \colon W_n \to [-1,1]$ be $(d_i|r_i)$-representable for $1 \leq i \leq m$ and $\lambda_1,\dots,\lambda_m \in \mathbb [-q,q]$. Then, the function
$$g = \beta(\lambda_1 f_1 + \cdots + \lambda_m f_m) \colon W_n \to [-1,1]$$
is $(d_1+\dots+d_m| 1+ \max_i r_i)$-representable.
\end{lemma}

The space of functions is naturally endowed with a normalized $L^1$-distance
$$\sigma(f,g) = \int_{W_n} |f(w)-g(w)| d\mu(w),$$
where $\mu$ denotes the normalized Lebesgue measure on $W_n$ 

For fixed $n,d,r$, the set of $(d|r)$-representable functions might be small, but nevertheless, we can use it to define alternative notions of distance on the set of all real-valued, measurable and essentially bounded functions on $W_n$ as follows:
$$\sigma_{(d|r)}(f,g) = \sup_h \left|\int_{w \in W_n} h(w)(f(w) - g(w)) d\mu(w) \right|,$$
where the supremum runs over all functions $h \colon W_n \to [-1,1]$ which are $(d|r)$-representable. It is clear that $\sigma_{(d|r)}$ takes non-negative values, that $\sigma_{(d|r)}(f,g) \leq 2 \sigma(f,g)$ and that $\sigma_{(d|r)}$ satisfies the triangle inequality. The first non-trivial observation is that $\sigma_{(d|r)}$ is actually a metric for any $d \geq 1, r \geq 0$.

\begin{lemma}
If $\sigma_{(d|r)}(f,g)=0$, then $f=g$.
\end{lemma}

The metric $\sigma_{(d|r)}$ measures how well $(d|r)$-representable  functions are able to tell the difference between $f$ and $g$. Note that it is very natural to include the complexity of the observer in any attempt of approximation of  functions by neural networks. We call a function $f \colon W_n \to [-1,1]$ $(\varepsilon,d|r)$-invisible if 
$$\sigma_{(d|r)}(f,0) = \sup_h \left|\int_{w \in W_n} h(w)f(w) d\mu(w) \right| \leq \varepsilon,$$
i.e., if it does not significantly correlate with any $(d|r)$-representable function.

We set
$$F(d,r) = \{f \colon W_n \to [-1,1] \mid f \mbox{ is $(d| r)$-representable} \}.$$
The following result is already interesting for $d=1,r=0$.

\begin{theorem}
Let $n \in \mathbb N$, $f \colon W_n \to [-1,1]$ be a measurable function, $d,r \in \mathbb N$ and $\varepsilon>0$. For $m = \lceil 1/\varepsilon^2 \rceil$, there exists 
a  function $g \colon W_n \to [-1,1]$ such that 
\begin{enumerate}
\item $\sigma_{(d|r)}(f,g) \leq \varepsilon,$ and
\item $g$ is $(2^m d,r+m)$-representable.
\end{enumerate}
In particular, every measurable function $f\colon W_n \to [-1,1]$ is a sum $f=g+h$, where
$g \colon W_n \to [-1,1]$ is $(2^m d,r+m)$-representable and $h=f-g$ is $(\varepsilon,d|r)$-invisible.
\end{theorem}
The proof is inspired by various analytic approaches to Szemer\'edi's Regularity Lemma, see for example \cite{MR2306658}. Note that our bounds are independent of $f$ and $n$, which should make the results particularly useful.
\begin{proof}
Consider the Hilbert space $L^2(W_n,\mu)$ with the usual inner product and consider a  function $f \colon W_n \to [-1,1]$ as a vector in $L^2(W_n,\mu)$. For $1 \leq k \leq m+1$, we set 
$$\Xi_k := \sum_{j=1}^k [-q,q] \cdot F(2^{j-1} d,r+j-1)$$ and define
$t_k := \inf \{ \|f - \xi\|^2 \mid \xi \in \Xi_k \}.$
We clearly have that $1 \geq \|f\|^2 \geq t_1 \geq \cdots \geq 0.$ Hence, there exists some $m' \leq m=\lceil 1/\varepsilon^2 \rceil$ such that
$t_{m'} \leq t_{m'+1} + \varepsilon^2.$ We conclude that there exists $g' = \lambda_1f_1 + \cdots +\lambda_{m'} f_{m'}\in \Xi_{m'}$ with the property that
$\|f-g'\|^2 \leq t_{m'+1} + \varepsilon^2.$
We consider now the vector $\xi:=(\lambda_1,\dots,\lambda_{m'}) \in \mathbb [-q,q]^{m'}$ and set
$$g := \varphi_\xi(f_1,\dots,f_{m'}) \colon W_n \to [-1,1].$$ Note that $g$ is just the composition of $g'$ with $\beta$ and hence $|f(w)-g'(w)|\geq |f(w) - g(w)|$ for all $w \in W_n$ which immediately implies $$\|f-g\|^2 \leq t_{m'+1} + \varepsilon^2.$$

Note that $f_j$ is assumed to be $(2^{j-1} d |r+j-1)$-representable for each $1 \leq j \leq m'$, so that we conclude by Lemma \ref{para} that $g$ is $(2^{m'}d,r+m')$-representable. Let now $h$ be any $(d|r)$-representable  function. Note that $g+th \in \Xi_{m'+1}$, for $t \in [-q,q]$, so that we have
$$\|f-(g+th)\|^2 \geq t_{m'+1} \geq \|f-g\|^2 - \varepsilon^2$$ and hence
$$t^2 \|h\|^2 + 2t \langle h, f-g \rangle + \varepsilon^2 \geq 0, \quad \forall t \in [-q,q].$$
If $\|h\| \geq \varepsilon \geq \varepsilon/q$, then we set $t= \pm \varepsilon/\|h\| \in [-q,q]$ and can conclude that $|\langle h,f-g\rangle| \leq \varepsilon.$ In the other case, when $\|h\| \leq \varepsilon$, we get more easily $|\langle h,f-g \rangle| \leq \varepsilon$ just using $\|f-g\| \leq 1$. This finishes the proof, since now
$$\sigma_{(d|r)}(f,g) = \sup_h |\langle h,f-g \rangle| \leq \varepsilon,$$
where the supremum runs over all $(d|r)$-representable functions.
\end{proof}

\section*{Acknowledgments}
This research was supported by ERC Consolidator Grant No.\ 681207. I thank Nihat Ay for interesting comments on a previous version of this preprint.

\end{document}